# Deep-seismic-prior-based reconstruction of seismic data using convolutional neural networks

Qun Liu, Lihua Fu, and Meng Zhang

*Abstract*— Reconstruction of seismic data with missing traces is a long-standing issue in seismic data processing. In recent years, rank reduction operations are being commonly utilized to overcome this problem, which require the rank of seismic data to be a prior. However, the rank of field data is unknown; usually it requires much time to manually adjust the rank and just obtain an approximated rank. Methods based on deep learning require very large datasets for training; however acquiring large datasets is difficult owing to physical or financial constraints in practice. Therefore, in this work, we developed a novel method based on unsupervised learning using the intrinsic properties of a convolutional neural network known as U-net, without training datasets. Only one undersampled seismic data was needed, and the deep seismic prior of the input data could be exploited by the network itself, thus making the reconstruction convenient. Furthermore, this method can handle both irregular and regular seismic data. Synthetic and field data were tested to assess the performance of the proposed algorithm (DSPRecon algorithm); the advantages of using our method were evaluated by comparing it with the singular spectrum analysis (SSA) method for irregular data reconstruction and de-aliased Cadzow method for regular data reconstruction. Experimental results showed that our method provided better reconstruction performance than the SSA or Cadzow methods. The recovered signal-to-noise ratios (SNRs) were 32.68 dB and 19.11 dB for the DSPRecon and SSA algorithms, respectively. Those for the DSPRecon and Cadzow methods were 35.91 dB and 15.32 dB, respectively.

*Index Terms*— convolutional neural networks, deep seismic prior, encoder–decoder, seismic data reconstruction

## I. INTRODUCTION

The collection of seismic data provides effective geophysical information for detecting geological formations and stratigraphic distribution. The quality of the data obtained from the receiver will directly affect subsequent seismic processing and interpretation, such as migration imaging, full waveform inversion, and so on. Owing to irregular or regular sampling of seismic data in space, the lack of signals will increase the difficulty of seismic data processing. Therefore, it is necessary to reconstruct seismic data so that subsequent data processing can better characterize complex geological structures.

There are different types of reconstruction methods, such as mathematical transform methods [1-3], prediction filter methods [4-6], and rank reduction methods [7-9]. The mathematical transform methods classify data into proper domains, such as the Fourier, curvelet, Radon, and shearlet domains, and may fail to reconstruct aliased seismic data owing to the sensitivity of parameter selection. The prediction filter methods assume seismic data to be a local superposition of linear events in the f–x domain, which need use a local window to guarantee linearity. The rank reduction methods assume that the Hankel matrix constructed using frequency slices or seismic data in a texture-patch arrangement is of a low rank, while noise or missing traces increase the rank of the data [7, 10]. Singular spectrum analysis (SSA) [11] is a classical method for seismic data reconstruction using rank reduction operations. This method estimates the real rank of seismic data; it is difficult to find an optimal rank; it usually performs well in case of irregularly missing traces and not suitable for regularly missing traces. Naghizadeh and Sacchi [12] proposed the de-aliased Cadzow method to reconstruct seismic data with regularly missing traces.

With the development of artificial intelligence, deep learning methods have attracted considerable attention. Generative adversarial networks have been explored for seismic noise attenuation and irregular trace reconstruction [13]. The training labels of neural networks are synthetic shots with real data geometry generated by a finite difference modeling engine; this would lead to unsatisfactory reconstruction quality because the labels are synthetic and not real. Wang *et al*. proposed a deep-learning-based approach that uses trained ResNets to reconstruct regularly missing traces. This method would appear biased when the feature in the test data is different from that in the training data and the bias increases as the differences increase [14]. Siahkoohi *et al*. [15] used generative adversarial networks to reconstruct seismic data regardless of the type of sampling. It can handle heavily undersampled seismic data; however, it requires that a percentage of shots fully sampled be available as training data. Oliveira *et al*. used conditional generative adversarial networks to reconstruct irregularly missing traces, which may present a lower resolution and some nongeological artifacts when the structure of data is complex [16]. A convolutional autoencoder network is proposed for seismic data reconstruction [17]. All the deep-learning-based methods mentioned above are under the framework of "training-validating-testing" and are invariably trained using

This study is financially supported by the National Key R&D Program of China (No. 2018YFC1503705), Science and technology research project of Hubei Provincial Department of Education (B2017597), Hubei Subsurface Multi-scale Imaging Key Laboratory (China University of Geosciences) under grants (SMIL-2018-06) and the Fundamental Research Funds for the Central Universities under Grants CCNU19TS020.

Q. Liu is with the Institute of School of Mathematics and Physics, China University of Geosciences, Wuhan 430074, China (e-mail: q.liu@ cug.edu.cn).
L. Fu is with the Institute of School of Mathematics and Physics, China University of Geosciences, Wuhan 430074, China (e-mail: lihuafu@ cug.edu.cn).
M. Zhang is with the School of Computer Science, Central China Normal University, Wuhan 430079, China (e-mail: m.zhang@mail.ccnu.edu.cn).

large datasets. However, it is difficult to acquire enough datasets of seismic data.

Therefore, in this study, a novel strategy for seismic data reconstruction based on deep seismic prior was developed and implemented using data with both irregularly and regularly missing traces. State-of-the-art deep learning methods are mostly trained using large seismic datasets; their excellent performance may be attributed to their ability to learn seismic priors from large datasets. Ulyanov *et al.* [18] first showed that convolutional neural networks (CNNs) have the intrinsic ability to handle ill-posed inverse problems without pre-training; they exploited the deep image prior using the inner structure of the CNN itself. Besides, seismic data have texture-patch structures and the neighboring patches are highly similar, which cause the CNN to extract deep seismic prior. CNNs with architecture similar to U-net have shown excellent performances in image inpainting [19], denoising [20] and super-resolution [21]. Inspired by the outstanding performance achieved in image processing problems, we exploited U-net [22] to capture deep seismic prior. Instead of following the common framework of "training-validating-testing" for a large seismic dataset, we fitted the U-net network as a generator to single undersampled seismic data. In this framework, the network weights were viewed as a parameterization of the model and were are randomly initialized. The seismic data reconstruction problem was recast as a conditional seismic generation problem, and the only information needed was contained in the input data and handcrafted structure of the network. Synthetic and field examples are provided showing promising performance in reconstructing data having either irregularly or regularly missing traces.

## II. Theory

### A. Problem formulation using CNN

In the reconstruction problem, $M \in \mathbb{R}^{m \times n}$ represents the observed seismic data that is undersampled (i.e., containing missing traces), and $X \in \mathbb{R}^{m \times n}$ represents the original complete seismic data. Further, $P_\Omega : \mathbb{R}^{m \times n} \to \mathbb{R}^{m \times n}$ is the masking operator, which is defined as

$$\left[P_\Omega(X)\right]_{ij} = \begin{cases} X_{ij}, & (i,j) \in \Omega \\ 0, & \text{otherwise.} \end{cases} \quad (1)$$

$\Omega$ denotes an index subset corresponding to observed entries. $P_\Omega$ is a projection on the observed data while keeping the entries in $\Omega$ and zeros at the locations of missing traces. The CNN was applied to generate missing traces using learning generator networks $X = f_\theta(Z)$; these map input noise $Z$ to seismic data $X$, where $Z$ is filled with uniform noise between 0 and 0.1 and has the same size as $X$; $f$ represents the network and $\theta$ represents network parameters. The original data $X$ can be reconstructed from a fixed input $Z$ filled with noise in a random distribution, where the distribution is conditioned on a corrupted observation $M$. For the reconstruction task, only missing traces had to be reconstructed by the network whereas the nonmissing traces remained untouched. Therefore, the loss function using the CNN was formulated as

$$\theta^* = \min_\theta \left\| P_\Omega(f_\theta(Z)) - P_\Omega(M) \right\|_F^2 \quad (2)$$

and

$$X = f_{\theta^*}(Z), \, s.t. P_\Omega(X) = P_\Omega(M). \quad (3)$$

$\theta^*$ is the optimal parameter of the network obtained by minimizing the loss function. $\left\| \cdot \right\|_F^2$ is the Frobenius norm, defined as $\left\| X \right\|_F^2 = \left( \sum_{ij} \left| X_{ij} \right|^2 \right)^{1/2}$.

### B. U-net structure for reconstruction

The U-net network used in this study has an encoder-decoder structure with skip-connections [23]. Fig. 1 shows the CNN-based network structure for seismic data reconstruction. It has the following features:

1) Convolution with a filter size of $3 \times 3$, followed by batch normalization (BN) [24] and Leaky rectified linear unit (LeakyReLU) [25]; these operations help the hidden representation. The numbers of filters used in this work were 16, 32, 64, 128, and 128, going deep into the network of the left-side encoder path.

2) Down-sampling by convolution with a filter size of $3 \times 3$ and stride of $2 \times 2$, instead of using max pooling to construct a full CNN. BN and LeakyReLU were employed.

3) Up-sampling through bilinear interpolation [26] with a filter size $3 \times 3$. The numbers of filters were 128, 128, 64, 32, and 16, moving up in the right-side decoder path.

4) Skip connections were formed by copying features from the left-side encoder path and adding them to the right-side decoder path, thus varying a small number of hyperparameters.

Note that the height and width of the input can be randomly chosen as needed. Noise-based regularization was used when fitting the network; that is, at each iteration, an additive normal noise was used to perturb the input $Z$. This regularization can help to better fit the network and achieve the desired objective. An ADAM optimizer [27] was used in experiments.

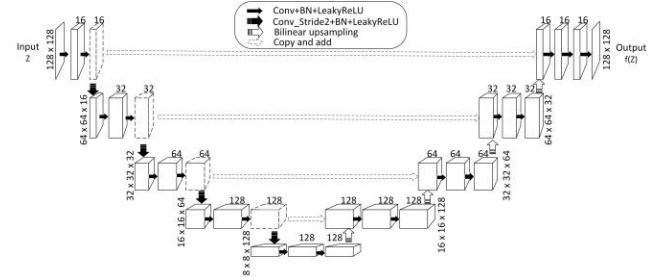

Fig. 1. Illustration of the U-net structure used in the experiments.

### C. Deep seismic prior

To show the intrinsic ability for exploiting deep seismic prior by the inner structure of the CNN network itself, an irregular reconstruction experiment using field data was implemented under different iterations using the deep-seismic prior-based reconstruction (DSPRecon) method, as shown in Fig. 2. In Fig. 2, we can observe that the CNN network using the U-net structure acts as a generator. When the number of iterations reaches 170, the textured structure of seismic data appears and it becomes clearer with increasing numbers of iterations.

To further show the deep seismic prior explored using the CNN, we developed feature maps of the layers in the U-net networks extracted from the output after 3000 iterations, as shown in Fig. 3. Owing to limitations in terms of the paper length, only some layers were chosen and the first few feature maps of chose layers are shown. Fig. 3 indicates that the deep seismic prior is encoded in down-sampling and well decoded in up-sampling, and the deeper layers extract deeper seismic prior. Fig. 4 shows the feature maps of the chosen layers under different iterations. It is clear that the shallow layers do not have obvious changes under different iteration numbers, while the deep layers have distinctive features under different iterations. Additionally, with the iteration numbers increasing, the extraction of the deep seismic prior continuously improves.

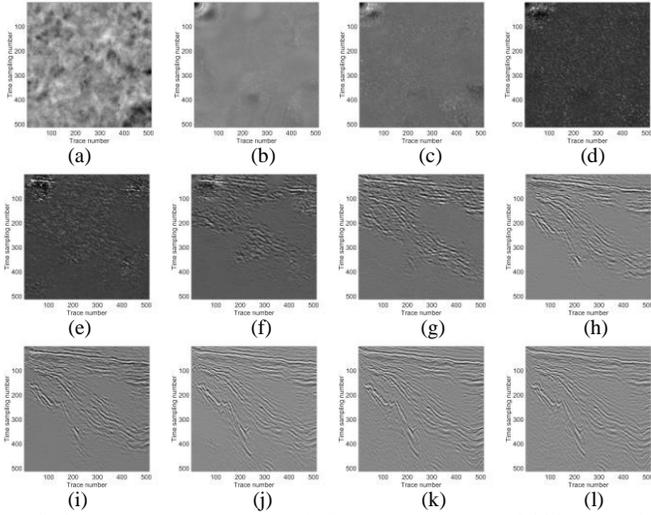

Fig.2. Reconstruction results under different iterations using DSPRecon method. (a) input noise, (b)–(k): the reconstruction results under different iterations, with the number of iterations being 50, 70, 100, 150, 170, 200, 240, 270, 400, and 600, respectively, and (l) output results (the number of iterations is 3000).

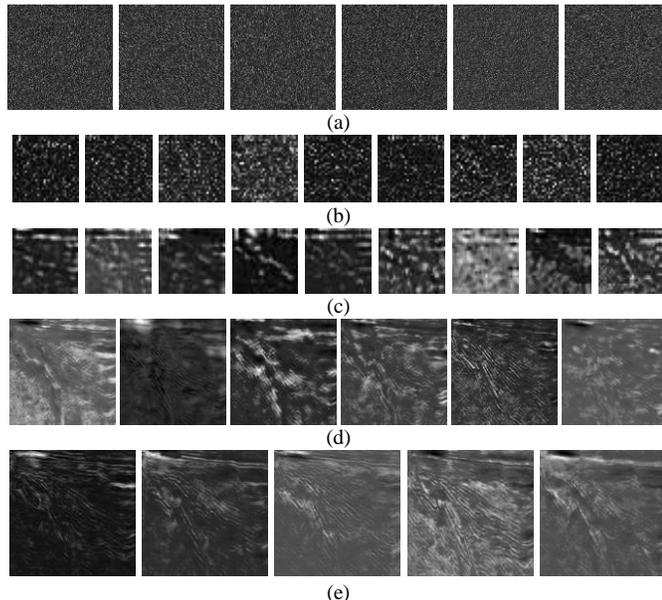

Fig.3. Feature maps showing different layers of U-net networks extracted from output after 3000 iterations. (a) first layer of down-sampling, (b) fourth layer of down-sampling, (c) first layer of up-sampling, (d) fourth layer of up-sampling, and (e) fifth layer of up-sampling.

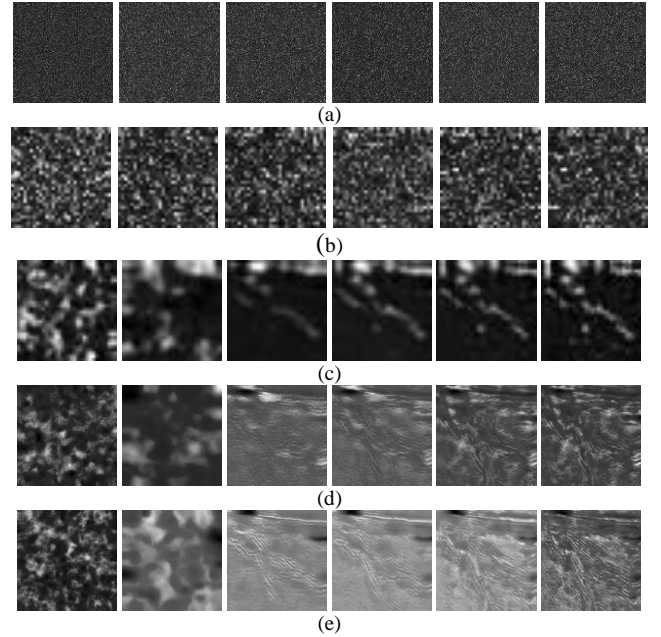

Fig. 4. The third feature map of each layer corresponding to Fig. 3 under different iterations with the number of iterations being 0, 70, 270, 400, 1000 and 3000. (a) the third feature map of the first layer in down-sampling, (b) the third feature map of the fourth layer in down-sampling, (c) the third feature map of the first layer in up-sampling, (d) the third feature map of the fourth layer in up-sampling, and (e) the third feature map of the fifth layer in up-sampling.

### D. Implementation details

We performed the experiments on a graphics processing unit that is NVIDIA GTX 1080-Ti, which took approximately 7 min. In our implementation, the seismic data were all normalized to the range 0–1. Input $Z$ of the network is uniform noise between 0 and 0.1 with a shape identical to the shape of processed seismic data. The number of iterations was set as 3000 in all experiments, and the learning rate was 0.001. At each iteration, input $Z$ was additionally perturbed with Gaussian noise of a specified variance, where we set $\sigma=0.03$. For reconstruction of irregularly missing traces, results were compared to the classical seismic data reconstruction method, the SSA method. For reconstruction of regularly missing traces, the results were compared to the de-aliased Cadzow method.

## III. EXPERIMENTS

The performance of our method was evaluated in both irregular and regular reconstruction. The quality of reconstruction is indicated by the signal-to-noise (SNR), defined as follows:

$$\text{SNR}=10\log_{10}\left(\frac{\|X\|_F^2}{\|X-X^*\|_F^2}\right), \qquad (4)$$

where $X$ and $X^*$ respectively denote the original seismic data and reconstructed data.

### A. Reconstruction of irregularly missing traces

One synthetic single-shot pre-stack seismic data with three curve events shown in Fig. 5(a) was tested; the number of time sampling points along the temporal axis is 256, and there are 256 traces along the spatial axis. The recovered SNRs were 32.68 dB and 19.11 dB for the DSPRecon algorithm and SSA algorithm, respectively, as shown in Figs. 5(c) and 5(e). Comparing Figs. 5(d) and 5(f) shows that the DSPRecon method provides better reconstruction performance. Fig. 6 presents the comparison of the *f–k* spectra. From Fig. 6(c), we can see that the DSPRecon method is effective for removing spatial aliasing. Fig. 6(d) shows that some artifacts still exist when using the SSA method. Fig. 7 shows the results of the reconstructed SNR and time consumption under different learning rates. It is evident that the reconstructed SNR tends to be stable when the number of iterations is 3000; moreover, a higher SNR could be obtained with learning rates of 0.01 or 0.001, and this was observed in all the experiments.

We tested our method on post-stack seismic data with a size of $512 \times 512$, as shown in Fig. 8. The recovered SNR was 33.09 dB using DSPRecon and 24.27 dB using SSA. Fig. 9 shows the reconstructed results of different single traces; it clearly shows that both the single traces reconstructed by DSPRecon fit the original signal well, and the reconstructed results are better than those produced by SSA. Fig. 10 presents the *f–k* spectra comparison of post-stack seismic data between DSPRecon and SSA. Fig.10(c) shows that spatial artifacts are all removed using DSPRecon, while Fig. 10(d) shows that the dealiasing effect when using SSA is not satisfactory.

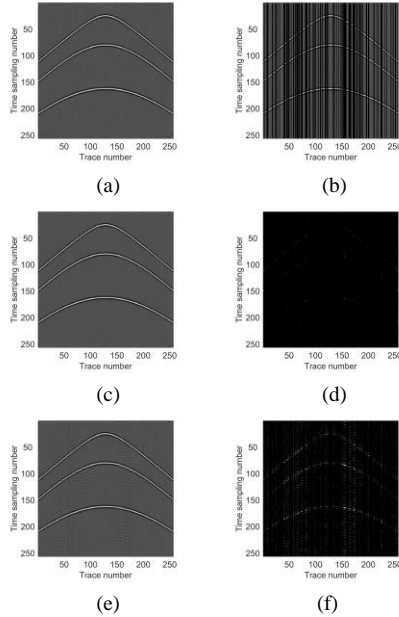

Fig. 5. Example of synthetic single-shot pre-stack seismic data reconstruction. (a) original data, (b) corrupted data with 50% randomly missing traces, (c) recovered data using DSPRecon, (d) residual between (a) and (c), (e) recovered data using SSA, and (f) residual between (a) and (e).

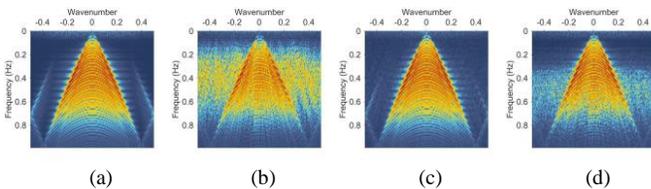

Fig. 6. *F–k* spectra comparisons of synthetic single-shot pre-stack data. (a) *f–k* spectra of complete original data, (b) *f–k* spectra of corrupted data with 50% randomly missing traces, (c) f–k spectra of recovered data using DSPRecon, and (d) f–k spectra of recovered data using SSA.

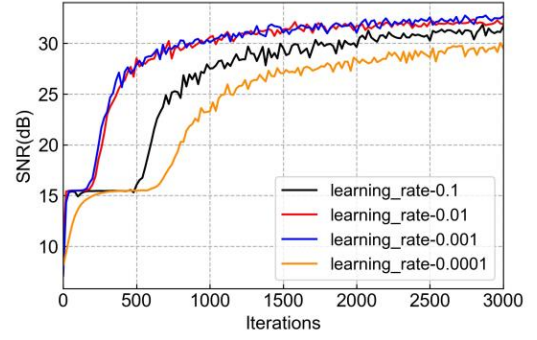

Fig.7. Reconstructed SNRs of synthetic single-shot pre-stack data during different iterations at different learning rates.

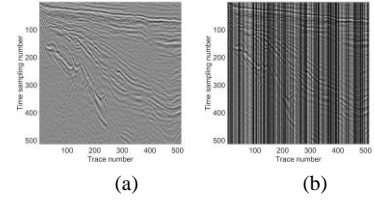

Fig. 8. Reconstruction of post-stack seismic data. (a) original data and (b) corrupted data with 50% randomly missing traces.

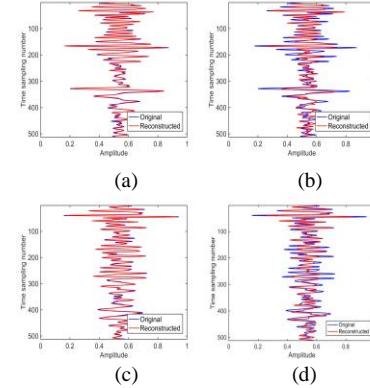

Fig. 9. Comparison of the 204th and 251st single traces taken from original post-stack data and reconstructed data. (a) and (b): the 204th single trace reconstructed using DSPRecon and SSA, respectively; (c) and (d): the 251th single trace reconstructed using DSPRecon and SSA, respectively.

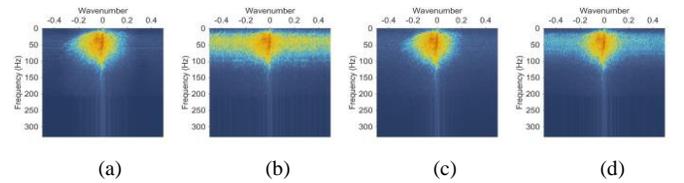

Fig. 10. *F–k* spectra comparisons of post-stack seismic data. (a) *f–k* spectra of complete original data, (b) *f–k* spectra of corrupted data with 50% randomly missing traces, (c) *f–k* spectra of recovered data using DSPRecon, and (d) *f–k* spectra of recovered data using SSA.

### B. Reconstruction of regularly missing traces

Field data with a size of $512 \times 512$ were tested to assess the applicability of the DSPRecon method. Fig. 11(a) presents one field data, including 512 traces and 512 time sampling points per trace. Fig. 11(b) presents the regularly sampled data with all the odd traces missing. The recovered SNRs using the DSPRecon and de-aliased Cadzow methods are 35.91 dB and 15.32 dB, respectively, as shown in Figs. 11(c) and 11(d). The

$f$–$k$ spectra are provided in Fig.12. Figs. 12(c) and 12(d) represent the $f$–$k$ spectra of the reconstructed field data using the DSPRecon and de-aliased CAdzow methods, respectively. Fig.12(c) is consistent with 12(a), and some artifacts exists in 12(d).

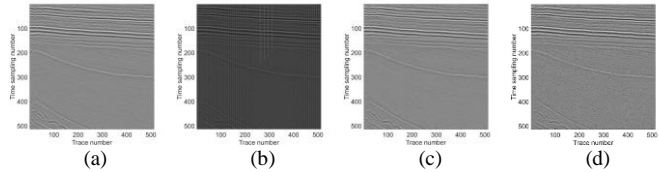

Fig. 11. Reconstruction of post-stack seismic data. (a) original data, (b) regularly sampled data with 50% missing traces, (c) recovered data using DSPRecon, and (d) recovered data using SSA.

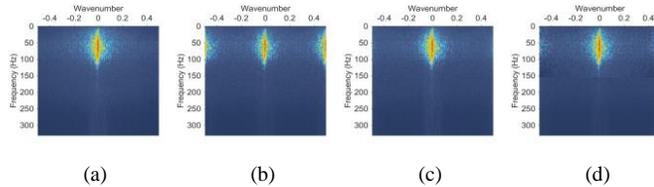

Fig. 12. $F$–$k$ spectra comparisons of post-stack seismic data. (a) $f$–$k$ spectra of complete original data, (b) $f$–$k$ spectra of corrupted data with 50% regularly missing traces, (c) $f$–$k$ spectra of recovered data using DSPRecon, and (d) $f$–$k$ spectra of recovered data using de-aliased Cadzow method.

## IV. CONCLUSION

Seismic data reconstruction was achieved for both irregularly and regularly sampled seismic data using a CNN based on deep seismic prior. First, the U-net network was designed and the experimental parameters were set. Then, a series of experiments on both irregular and regular reconstruction were performed. The reconstructed results showed the flexibility and validity of the proposed method. The DSPRecon method differs from traditional deep learning methods that need large datasets for training; it only depends on the corrupted data and thus saves costs associated with acquiring datasets and time consumption in training. Moreover, it uses deep seismic prior instead of rank prior compared to rank reduction methods, avoiding the choice of rank. As satisfactory performances are shown in seismic data reconstruction using the DSPRecon method, we intend to extend this method for denoising seismic data.